# Assessing robustness of machine learning models using covariate perturbations

Arun Prakash R[1], Anwesha Bhattacharyya, Joel Vaughan, and Vijayan N. Nair

Model Risk Management, Wells Fargo[2]

May 2024

## Abstract

As machine learning models become increasingly prevalent in critical decision-making models and systems in fields like finance, healthcare, etc., ensuring their robustness against adversarial attacks and changes in the input data is paramount, especially in cases where models potentially overfit. This paper proposes a comprehensive framework for assessing the robustness of machine learning models through covariate perturbation techniques. We explore various perturbation strategies to assess robustness and examine their impact on model predictions, including separate strategies for numeric and non-numeric variables, summaries of perturbations to assess and compare model robustness across different scenarios, and local robustness diagnosis to identify any regions in the data where a model is particularly unstable. Through empirical studies on real world dataset, we demonstrate the effectiveness of our approach in comparing robustness across models, identifying the instabilities in the model, and enhancing model robustness.

## 1  Introduction

The term "robustness" has a long history in statistics and data science. Early usage was concerned with need for inferential methods that were not overly sensitive to the presence of outliers in the data. (Box 1979) appears to be the first to discuss the notion of model robustness. His definition of robustness "as the property of a procedure which renders the answers it gives insensitive to departures, of a kind which occur in practice, from ideal assumptions" is very general and covers different types of robustness. Since Box's pioneering paper, there has been extensive work in the statistical literature on model robustness. Early work dealt with parametric models, where one examines the behavior of procedures under model misspecifications. With increasingly large sample sizes, advances in computing, and renewed interest in flexible nonparametric models (using machine learning algorithms), we were able to relax restrictive parametric assumptions and fit increasingly complex models. These led to the opposite problem where the models become too flexible and can overfit the training data. One way to address this in the model fitting stage is through regularization, and there is a vast literature on regularization techniques and the associated hyperparameter tuning algorithms. Despite these approaches, there is still a need for methods to assess the robustness of a model after it has been fit to the training data. This is the focus of the present paper.

The main reason for assessing model robustness is the generalizability of a model on unseen data. A common way to measure this is by comparing predictive performances between train and test datasets (which are typically hold-out data sets, in-time, out-of-time test sets, or subset of the entire dataset). An overfit model fits to the noise in the training data and thus, despite producing low training error, but it can

---

have a comparatively large test error leading to poor performance when used in production, as discussed in (Groh 2022). The large difference between training and test datasets is often called gap. To understand the reasons and regions where the over-fit occurs, it is useful to segment the test and training datasets into natural groups (by predictors or regions such as tails, areas with sparse data, etc.) and compare the gaps. Such comparisons provide useful insights that can be exploited to improve the model fit on the test dataset. This approach is examined in detail elsewhere in our work.

Another technique to assess model robustness is by examining its complexity: more complex models such as machine learning algorithms tend to learn the bumps in the training datasets that do not necessarily generalize to test datasets. In linear models, model complexity is captured through the notion of degrees-of-freedom. There are papers that generalize this concept to flexible nonparametric models, including machine learning algorithms. We have examined the use of the generalized degrees-of-freedom and how they can be used to assess robustness. These results will be reported in another paper.

This paper focuses on techniques to assess robustness by perturbing the covariates or predictors and examining the reduction in predictive performance. Before moving to the main part of the paper, we should note that "model robustness", as used currently in the literature, is a broad term that encompasses performance of the model on shifting data distributions that might arise from covariate shift (change in the distribution of predictors), prior probability shift (change in the distribution of the response), concept shift (change in the conditional distribution of the response given covariates), etc. Each of these is a hard problem, and one cannot expect to find a model that is "robust" to all these changes. However, in dynamic industries and evolving environments like finance, supply chain, etc., new data distributions arise, and these can differ from historical distributions that were used to train the models; thus, it is necessary to anticipate the risk in prediction in scenarios where the distributions evolve. The paper by (Wu, et al. 2023) presents a framework for certifying robustness against real-world distribution shifts for deep neural networks. The paper (Taori, et al. 2020) focuses on how robust ImageNet models are to distribution shifts arising from natural variations in the data. The idea of adversarial robustness has gained traction, and researchers have studies the robustness of models against adversarial attacks (see, for example, [ (Mohus and Li 2023), (Ruan, Yi and Huang 2021)].

The approach taken in this paper is somewhat similar to the adversarial literature although we examine changes in performance to small perturbations in the covariates/predictors rather than huge changes. This involves:
a) Defining the notion of "small" perturbations of covariates and deciding how to perturb them, and
b) Summarizing the results of perturbations.

Assessing model robustness is easier when one compares multiple models (or algorithms). It is more difficult to determine robustness in isolation. In addition, if the model is found to lack robustness and it is the recommended model by stakeholders, one may want to examine the reasons and, if needed, find ways to mitigate the problem.

We conclude this section with a brief outline of the remainder of the paper. In section 2, we discuss our approach and the underlying methodology to assess the robustness of models. We also introduce the robustness metric, discuss different metrics to summarize the perturbations, and the effect of the perturbation budget on the robustness metric. In section 3, we provide details of the perturbation strategy for numeric and non-numeric variables. In section 4, we introduce local diagnosis tools to identify the variables and regions in data that are contributing to the volatility in predictions. In section 5, we illustrate the methodology on a public dataset (Yeh 2016) by comparing multiple models on their robustness. We demonstrate the effect of budget on the robustness metric, different perturbation strategies for numeric and categorical variables, and the usage of local diagnosis tools. We conclude in section 6.

## 2 Covariate perturbations

### 2.1 Methodology

Let us define the dataset to be perturbed as $X^{n \times p}$ with $x_i$ denoting the $i_{th}$ observation vector. $y_i$ is the corresponding true response and $\hat{y}_i$ is the predicted value from a Model defined as $\hat{y}_i = Model(x_i)$.

The underlying approach is as follows:

1. For each observation $x_i$, we perturb $x_i$ K times in a local neighborhood $LN(x_i, b)$, where $b$, the budget, is a measure of locality. If $b = 0$, there is no perturbation. We let $\Delta x_{ik}$ denote the k$^{th}$ perturbation on observation $x_i$ with perturbed observation being denoted as $x_i + \Delta x_{ik}$.
2. Calculate $\hat{y}_{ik} = Model(x_i + \Delta x_{ik})$
3. Summarize the variability in $\hat{y}_{ik}$ as a measure of robustness, as explained in 2.2.

### 2.2 Summary of Perturbations

We summarize resulting predictions as follows:

For each perturbation $k$ of an observation $i$, we compute $\hat{y}_{ik} - \hat{y}_i$, the difference between the model prediction on perturbed data point and model prediction on original data point. Next, we summarize the deviances for an observation $i$ using root mean square across all K perturbations, thus giving us the root Perturbed Prediction Volatility (rPPV) for observation $i$,

$$rPPV_i = RMS(\hat{y}_{ik} - \hat{y}_i) = \sqrt{\frac{\sum_{k=1}^{K}(\hat{y}_{ik} - \hat{y}_i)^2}{K}}$$

We then compute the average of $rPPV_i$ across all $n$ observations. We define this metric as Average root Perturbed Prediction Volatility (ArPPV), a measure of robustness.

$$ArPPV = \frac{1}{n}\sum_{i=1}^{n} rPPV_i$$

This aggregate measure can be used to compare multiple models on their local stability in predictions. It is to be noted here that the above robustness measure can be formulated using different summarization metrics. Each summarization metric will capture a different notion of robustness. For instance, one could use the mean square as a summarization metric instead of root mean square (Average Perturbed Prediction

Volatility (APPV)) in cases where a measure in the same scale as Mean Squared Error (MSE) is desired. Both of these formulations gives an average deviation across the K perturbed data points. Alternatively, one could use absolute maximum to estimate the largest change in prediction in the neighborhood among the K perturbed data points, providing a worst-case measure rather than an average view. Similarly, maximum square, absolute mean, absolute median, etc. are few other summarization metrics than can be used in different context. For the rest of the paper, we shall use ArPPV as the standard metric to capture the average change in prediction for multiple perturbations.

While summarizing the variability in $\hat{y}_{ik}$, it is possible to summarize the K perturbations by their deviation from the original prediction $(\hat{y}_{ik} - \hat{y}_i)$ or by their deviation from the true response $(\hat{y}_{ik} - y_i)$. We expect the model predictions to be stable for small-scale perturbations, so if on average, the deviation from original prediction $(\hat{y}_{ik} - \hat{y}_i)$ is high for an observation $i$, then it is likely that the prediction is not stable at that point. On the other hand, if on average the deviation from true response $(\hat{y}_{ik} - y_i)$ is high for an observation $i$, then this is a combination of model bias $(\hat{y}_i - y_i)$ and model stability $(\hat{y}_{ik} - \hat{y}_i)$. Thus, to understand the smoothness of the prediction surface, it is intuitive to look at $(\hat{y}_{ik} - \hat{y}_i)$. Also in cases when the response is binary and the problem is binary regression, then $\hat{y}_{ik}$ denotes the predicted probability of the perturbed point and other metrics would be necessary to incorporate loss in model performance; however, if we are only interested in a metric to define stability of predictions, then we can use $(\hat{y}_{ik} - \hat{y}_i)$, irrespective of response type. Note that, $(\hat{y}_{ik} - \hat{y}_i)$ may also be high due to sensitivity of the model to one or more variables at the locality of observation $i$, particularly if there exists a sharp but systematic transition in that neighborhood. In general, it is difficult to distinguish between model sensitivity and lack of stability, and a model demonstrating instability using this metric may need to be scanned for sensitivity issues in downstream analysis. We will discuss this further in section 4.

ArPPV increases with budget $b$, where budget is defined as the measure of locality around the data point that is perturbed. The budget should be kept small to measure the robustness of models. For small budgets, a robust model should give stable predictions, thus producing small ArPPV values, while for large budgets, the model is expected to be sensitive to large perturbations of important variables; otherwise, the model is flat and may be underfit. Since ArPPV is not bounded above, it is more useful for comparing multiple models than being used to assess a single model in an absolute sense.

## 3   Perturbation Strategy

In the following, we propose a strategy to generate perturbations that are local and try to maintain the data envelope and inherent associations in the data when possible. While the concept of a local neighborhood is well-defined in continuous or naturally ordered discrete variables, non-numeric or categorical variables represent a coarse segmentation of the data space, such as delinquency status of an account. Perturbing such variables from one state to the other creates a large disturbance in the system. Often the distributions of numeric variables are different in each segment and hence perturbing a categorical variable would imply that the other covariates be changed drastically as well to maintain association. Thus, perturbing such variables may not be appropriate for robustness tests in general. However, in certain circumstances, there might be a need to test the robustness of models against perturbations in these predictors (for example, a model where the majority of predictors are categorical). Keeping this in mind, we propose a method that generates perturbations of numeric variables independent of the non-numeric (categorical) variables.

For numeric variables, we can generate multiple random perturbations from a multivariate Gaussian distribution respecting the correlation structure of the data. If all observations are perturbed by similar amount, then we call this the raw perturbation strategy. On the other hand, a variable may not be uniformly distributed, and it may be desirable to perturb an observation lying in a dense data region less than observation lying in a sparse region. To achieve this, we can make the size of the perturbations adaptive to the variable's density. We call this the adaptive perturbation strategy.

For categorical variables, it is difficult to induce 'local' perturbations as there is no inherent concept of distance. We address this issue by introducing a concept of pseudo-distance between categorical observations and use this distance to create local perturbations. We also ensure that all categorical perturbations respect the data envelope by restricting the perturbations to configurations that are present in the training data.

## 3.1 Numeric Variables

We first discuss strategies for perturbing numeric variables.

### 3.1.1 Raw perturbations

For numeric variables, the perturbations are generated from a Gaussian distribution whose correlation structure is taken from the original data. Let the estimated Pearson correlation structure be $\hat{P}$, then for a given budget $b$, the $j_{th}$ component of $k_{th}$ perturbation of observation $i$ is obtained as:

$$\Delta x_{ikj} = \epsilon_{ikj} \times b \times \sigma_j \tag{1}$$

where,

$$\sigma_j : sd(X_j) \tag{2}$$

$$\epsilon_{ik} \sim N(0, \hat{P}) \tag{3}$$

Thus, the perturbed observation for $k_{th}$ perturbation of observation $i$ is given as:

$$\tilde{X}_{ik} = X_i + \Delta x_{ik} \tag{4}$$

Budget ($b$) controls the degree of perturbation around the observation and it is incorporated as a percentage of the standard deviation of each variable ($\sigma_j$). Hence for a variable $X_j$ with a unit variance, a 2% budget would imply a perturbation range of 0.06 around the original value with a probability of 99.7%. For a zero budget, data is not perturbed, resulting in zero ArPPV. For an increasing budget, the perturbations become larger and larger, resulting in a larger ArPPV.

Preserving the correlation in the perturbations ensures that we do not distort the data envelope. For instance, if we have a data with 'inflation rate' and 'unemployment rate' as two of the variables, increasing 'inflation rate' should result in decreased 'unemployment rate'. This pattern should be observed in perturbed data as well. However, for low budgets, ignoring the correlation is acceptable as its impact on data distortion is extremely low.[3]

---

[3] The correlated strategy is basically the same as adding noise to all the principal components and reverting to raw scale; the underlying assumption in both cases being that the association between the predictors are mostly linear.

The perturbations can also be independent across the variables, and this is achieved by using the identity matrix instead of $\hat{P}$ when generating the perturbations. It is expected that for very small budgets, correlated and independent perturbations are similar and cause no significant difference in the correlation structure of the perturbed data. As we increase the budget, the difference becomes non-trivial, as shown below in Figure 1. To respect the data envelope, we can modify the out-of-range perturbations to return to the minimum/maximum value of the variable in the original data.

### 3.1.2 Adaptive perturbations

In the above-described raw perturbation strategy, we perturb all observations to same noise scale $\sigma_j$ defined in equation (2). However, in most situations, there is considerable variation in the dataset locally. For example, data in the tails or in sparse regions usually have larger variation. Thus, intuitively one should take the differences in variation into account in the definition of local perturbations. To achieve this, we make the perturbations adaptive by adjusting the noise scale $\sigma_j$ to $\tilde{\sigma}_{ij}$ where each observation '$i$' has its own scale based on the spread of nearby points. Note that the direction of association (positive/negative) is still retained in $\epsilon_{ik}$ in equation (3) although the scales of perturbations change for each observation.

While there are several ways to get the local measure of spread for each observation, we propose bucketing each variable into quantile bins and computing a rolling mean of the standard deviation in each bucket. Let $s_q$ be this measure of the spread in bucket q. We adapt the scale of noise for observation '$i$' lying in bucket q as:

$$\tilde{\sigma}_{ij} = \frac{s_q}{\min(max(s_q), \sigma_j)} \sigma_j \quad (5)$$

If $s_q$ is very small compared to $max(s_q)$ or $\sigma_j$, then $\tilde{\sigma}_{ij}$ becomes smaller than the original scale of perturbations. If $s_q$ is greater than the global measure of spread, $\sigma_j$ then $\tilde{\sigma}_{ij}$ is $s_q$.

**Table 3-1. Comparison of variance of perturbations from raw and adaptive type for different variables. (Left): Distribution of variable, (Middle): Perturbation variance of each observation for raw perturbations, (Right): Perturbation variance of each observation for adaptive perturbations.**
**[Refer to Section 5 for description of variables and dataset]**

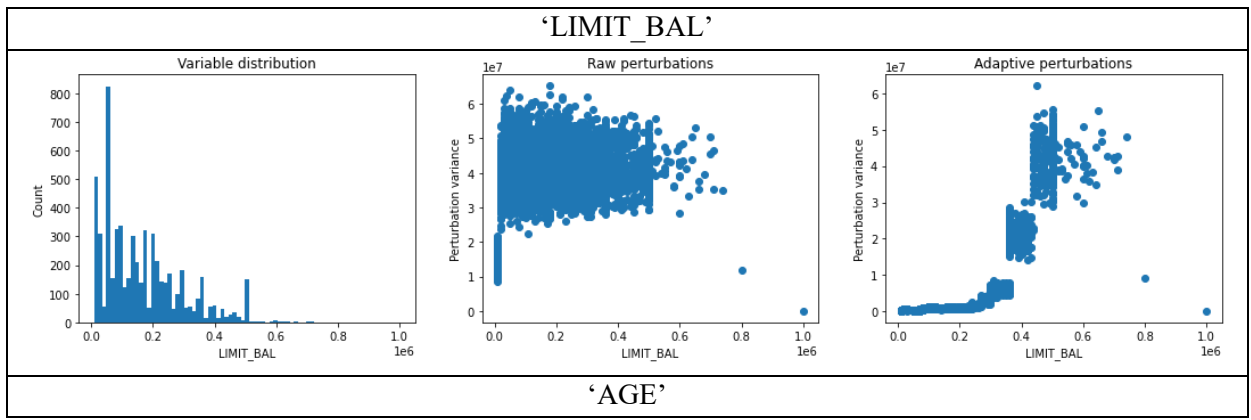

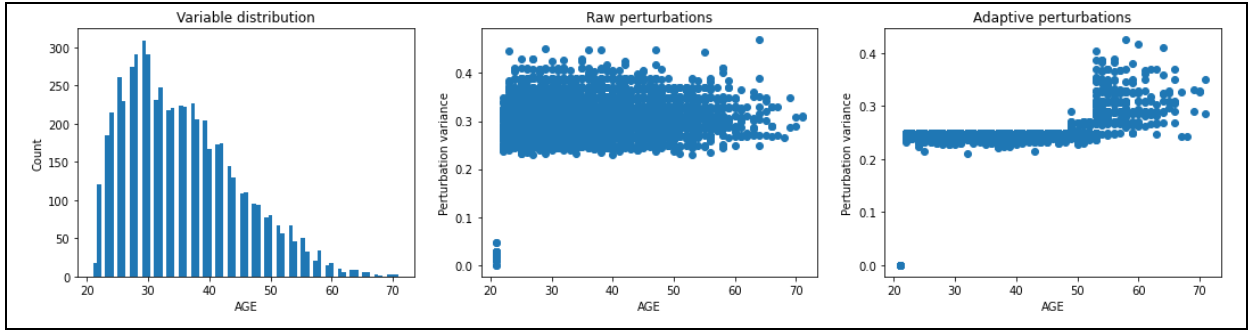

If a variable is uniformly distributed, we expect the adaptive perturbations to be similar to the raw perturbations as $s_q/max(s_q)$ is close to 1. When the distribution is non-uniform, we expect perturbations to be more conservative in dense region and relaxed in sparse/long-tailed regions. A comparison of the raw and adaptive perturbations is presented in Table 3-1. It shows that for raw perturbations observations have similar perturbation variance across the entire range of variables (except in tails where variance is reduced due to modification on out-of-range perturbations). For adaptive perturbations we can clearly see that in high density region, the perturbation variance is much lower compared to low-density regions resulting in over-all more conservative perturbations than the raw strategy. Thus, the two strategies work as intended.

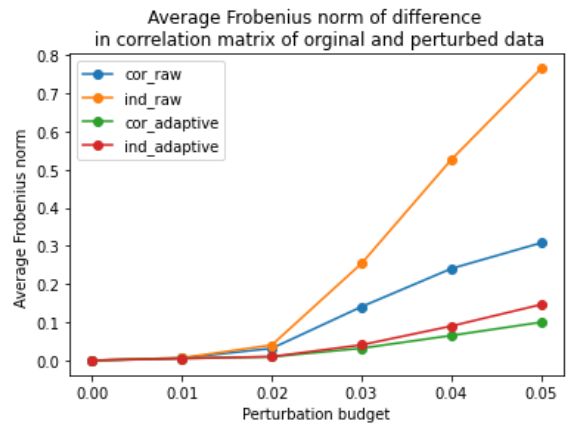

**Figure 1. Comparison of correlated and independent methods of perturbation for numeric variables by observing the average Frobenius norm of difference in correlation matrix of original and perturbed data.**

Figure 1 shows the average Frobenius norm of difference in the correlation matrices of original and perturbed data plotted against the perturbation budget. We observe that for small budgets the difference between the methods is insignificant and the norm is close to 0 implying correlation structure of perturbed data is close to original data. As budget increases, the average Frobenius norm of independent perturbations increases faster than correlated perturbations. The adaptive perturbations are usually more conservative, hence the perturbed data in this case is closer to original data than in the case of raw perturbations.

### 3.1.3 Perturbation of discrete variables

Once the data is perturbated, we round the perturbed values of discrete variables to their nearest integer value to maintain the data type. This ensures that we do not create perturbed values that are uninterpretable or invalid for such variables. The perturbed values are thus consistent with possible values for the variable and the nature of a model's interpolation between possible values does not become a factor in the robustness test. We don't want to penalize the algorithm for making a random decision in a space where realistically the variable will not occur. It is a necessary step to maintain consistency with the nature of the data.

Note that due to this rounding up, a discrete variable can only be perturbed if the noise to be added has absolute value $\geq 0.5$. Hence discrete variables may need either a large budget or a larger noise scale in order to be perturbed. In our applications in Section 115, we chose to artificially inflate the noise scale for discrete variables to produce reasonable perturbations at chosen budgets.

## 3.2 Categorical Variables

In general, it is less prudent to perturb categorical variables as they often represent coarse segmentation of the population, and such perturbations may not abide the notion of local perturbations. Especially perturbing categorical variables which have strong association with numeric predictors can distort the data envelope. For example, if we have 'season' and 'temperature' as categorical and numerical predictors respectively in a dataset, then perturbing 'season' will imply a large change in 'temperature'. Thus, one should be conscientious and deliberate when perturbing categorical variables.

Nevertheless, if the robustness test requires perturbation of categorical variables which are independent of other numeric variables in the data, we need a method for perturbing them independently. The main challenge lies in the fact that there is no sense of ordering or distance between unique values or levels of these variables.

A simple way to perturb categorical variables is to randomly shuffle the value of a variable using its marginal distribution. For the rest of the paper, we refer to this as the 'shuffling' strategy. However, there is no notion of locality (local perturbation) in this method. For instance, changing a categorical variable like 'Gender' from 'Male' to 'Female' may not be considered as a local perturbation because firstly it represents two very different segments of the population in multiple aspects and perturbing Gender may have a large impact on the response. Also, for variables with multiple levels in a categorical variable like 'weather-situation' in 'Bike Sharing' (Fanaee-T 2013) dataset, which may have levels like 'foggy', 'rainy', 'sunny', etc., changing or perturbing the 'weather-situation' from 'foggy' to 'rainy' is intuitively a smaller change than changing 'foggy' to 'sunny' in terms of the various attributes that define a weather situation (humidity, temperature, etc.). In such scenarios, the former change may be considered as a smaller perturbation in some sense. However, random shuffling does not consider this difference. Another drawback of random shuffling for categorical variables is that the association among variables is no longer maintained. The perturbed data may contain a combination of values for the categorical variables that do not make any logical sense and/or lie well outside of the data envelope.

In the following section, we propose a perturbation strategy based on a pseudo-measure of nearness between different levels of categorical variables based on their average impact on the response. In the following sections, we define the pseudo-distance and the perturbation strategy developed based on this distance.

### 3.2.1 Pseudo-distance measure

We want a distance measure that captures closeness between the levels of one or more categorical variable. This can be based on subject matter expertise or through careful scientific evaluation. However, without such information, we rely on the data to create a measure. We define a pseudo-distance between levels of categorical variables based on their average impact on the response. The underlying idea is that if two levels of a categorical variable are very similar then their average impact on the response should not be very different. Note that this is a data-driven approach to induce a distance metric on the levels of the categorical variables and may not necessarily capture the true closeness which a careful science-driven approach may.

For a given categorical variable $x$ with levels $l_1, l_2, \ldots, l_m$, we define a distance measure between any two levels $l_i$ and $l_j$ as the absolute difference in average response of observations at the two levels.

$$d(x = l_i, x = l_j) = \left| avg(y_{x=l_i}) - avg(y_{x=l_j}) \right| = d(x = l_j, x = l_i)$$

This measure is symmetric and defines a concept of similarity based on whether the two levels have similar or different average impact on the response. We can compute an $m * m$ distance matrix for each categorical variable with the cell $(i, j)$ given by $d(x = l_i, x = l_j)$.

We further Min-Max scale the distance matrix so that the most disparate levels have distance 1 in order to maintain consistency among different variables and keeping a single variable from dominating the categorical perturbations.

Now, for each categorical variable $x_j$, we have a corresponding distance matrix $d_j$. Thus, we can define the distance between two observations $x$ and $x'$ on $p$ categorical variables as:

$$D(x, x') = d_1(x_1, x_1') + d_2(x_2, x_2') + \cdots + d_p(x_p, x_p')$$

In some cases, we might want to control the number of perturbations of certain variables compared to others. We can modify the above formula to have that flexibility by giving weights to each variable.

$$D(x, x') = w_1 d_1(x_1, x_1') + w_2 d_2(x_2, x_2') + \cdots + w_p d_p(x_p, x_p')$$

The weights can be based on variable importance or any other criteria. The higher the weight on a variable, the lesser the chance of that variable getting perturbed. This may be used to ensure that the important model variables are not easily perturbed during the joint perturbation. Giving more weights to sensitive variables will produce more conservative perturbations.

### 3.2.2 Perturbation strategy

We propose joint perturbation of the categorical variables based on a single budget. This budget helps in controlling the number of variables that will change in each perturbation and the levels they can take. The strategy maintains associations among variables and ensures that perturbed points lie within the data envelope. For instance, if we have two variables, one being an indicator of 'working day' and other being the indicator of 'Sunday', then the 'Sunday' indicator would not be changed to 'non-Sunday' while keeping the 'working day' indicator as 'False' unless such combination exists in the data.

We first define $\mathcal{X}$ as the set of combinations of categorical observations present in the training data. This is referred to as the data envelope. Now, given a budget $b$, for each observation $x$, we define a subset of $\mathcal{X}$:

$$\mathcal{X}_x^b = \{z \in \mathcal{X};\ D(z,x) \leq b * \max(D) = b * p\}$$

This set now consists of combinations that are not too different from observation $x$ when measured with the distance measure $D$. The amount of dissimilarity is bounded by the budget $b$. While perturbing $x$, we want to restrict the perturbed values to this set. Also note that due to the Min-Max scaling of the distance matrices, $\max(D) = p$.

For generating $K$ perturbed versions of observation $x$, we sample $K$ times with replacement from the set $\mathcal{X}_x^b$. Finally, we accept each perturbation with a probability $max\_prop$ or reject the perturbation and retain the original observation. $max\_prop$ indicates the proportion of $K$ perturbations to be accepted for each observation.

The stochastic acceptance step ensures that we do not have cases where all the perturbations change the original value of a particular variable. This is especially helpful when we want to perturb a single binary variable. To perturb this variable, the budget must be increased to 100%, as none of the observations will get perturbed for any lower budget. But for 100% budget, all the perturbations will swap the variable's value, potentially leading to a substantial impact. Thus, the use of $max\_prop$ allows us to have flexibility on the proportion of perturbations per observation. We further illustrate this strategy and compare it to the random shuffling method in section 5.3.

Note that this strategy in essence induces a transition matrix for each observation $x$, which puts equal opportunity to be perturbed to all configurations in $\mathcal{X}_x^b$ and no opportunity to transition to configurations outside this set. As $b \to 0 \implies \mathcal{X}_x^b \to \phi$ and $b \to 1 \implies \mathcal{X}_x^b \to \mathcal{X}$. Thus, the pseudo-distance helps us set the transition probabilities based on a budget without having to explicitly define a (potentially large) transition matrix.

## 4 Local diagnosis

In addition to comparing models with respect to robustness, we can use the observation level measures of robustness defined in section 2.2 to understand the behavior of model locally and identify regions of the data where a model shows a lack of robustness. In this section we describe these local measures and illustrate their use.

As discussed in section 2, we create multiple random perturbations of each observation in a local neighborhood and summarize these perturbations to quantify the smoothness of the prediction surface around that observation. We have defined the observation level measure $rPPV$ in section 2.2, which summarizes the deviations of predictions at perturbed observations from prediction at the observed point of all perturbations of the given observation. We can then use this measure to identify the variables/regions contributing significantly to the volatility in the predictions. We describe two approaches in this section. The first uses the population stability index (PSI), a statistical measure that quantifies the difference between a probability distribution and a reference distribution. The second approach uses a supervised partitioning tree.

In the first approach, we choose a threshold to identify the observations with the highest rPPV measures, for example the highest 5% observations. We can then use PSI to compute the difference between the subset of observations with the highest rPPV and the rest of the observations for each variable. Ranking the variables by their corresponding PSI, we can identify the set of variables that display maximum shift in distribution. Further diagnosing these variables will help identify regions where the response is most volatile.

Alternatively, we create a diagnostic supervised partition tree on the covariate space using each observation's rPPV as the response. At each leaf node, we obtain the average rPPV for observations in that node. The variables of interest are those that have resulted in a node split, and particularly the nodes that result in the maximum difference between the average rPPV values of its leaf nodes. This set of variables identifies high volatility regions in the dataset and diagnosing these variables helps us understand the causes of the volatility in the predictions of the given model.

After identifying a set of variables, we investigate each variable separately by perturbing observations on that variable only keeping the values of other variables fixed. These single variable perturbations isolate the contribution of the selected variable to each observation's volatility, which can be visualized using a scatterplot of rPPV against the variable. This plot immediately shows the high volatility regions in this variable.

However, regions may have high rPPV due to appropriate model sensitivity rather than overfitting or problems with the model. For example, the model might be responding to interactions, transitions, or other variables, resulting in high rPPV in that region. Thus, comparing the scatterplot of rPPV vs variable with Partial Dependence Plot (PDP) [ (Friedman 2001), (Hu, et al. 2020)] of that variable helps us understand better the reason behind high rPPV regions. The PDP shows the marginal effect of a variable on the predicted outcome of a model and shows the relationship between the target response and variable. Thus, regions with high rPPV might be arising due to a transition effect in this relationship that can be observed in the PDP. The transition points and regions with large slopes on the PDP have a correspondingly high rPPV. If this transition is expected or reasonable, then the observed high ArPPV is caused by model sensitivity and should not be attributed to lack of robustness. However, if two models have similar performance and both have captured expected model activity but have different ArPPV, then the model with higher ArPPV can be deemed as less robust to small scale perturbations.

Another way to understand the behavior of the model in a local region is to check for monotonicity. As the generated perturbations are local, the model's prediction is expected to be monotonic in that small region. This can be measured by computing the number of times the prediction surface has a change in its second derivative. We expect a robust model to have a small number (0 or 1) of monotone violations. For example, a linear model will never have any monotonicity violations for local perturbations on a single variable. Capturing the number of monotone violations on the set of observations with high rPPV helps us measure how unstable the model is in those regions. If an observation with high rPPV has 0 monotone violation, then we can attribute the high value to model sensitivity in the region as opposed to model instability. It is to be noted that choice of variables, budget and quality of perturbations play a crucial role is assessing robustness to small scale addition of noise in data and these factors should always be carefully examined before arriving at a conclusion. We have further illustrated these local measures in section 5.4.

## 5 Illustrations

In this section, we illustrate the methodology to compare the robustness of three different models. We have used the (Yeh 2016) dataset that contains the credit card customer's default payment status in Taiwan (henceforth referred to as Taiwan credit dataset) along with other information like gender, marital status, educational qualifications, age. It also includes monthly information regarding amount of the given credit, history of past payments, amount of bill statement and previous payments for the period April to September 2005. The response variable is a binary indicator of default payment (1 = Yes, 0 = No). The other variables are described in Table 8-1. In Figure 2, we show the correlation heatmap, and observe that there are a few sets of highly correlated variables present in the dataset like 'PAY', 'BILL_AMT', etc.

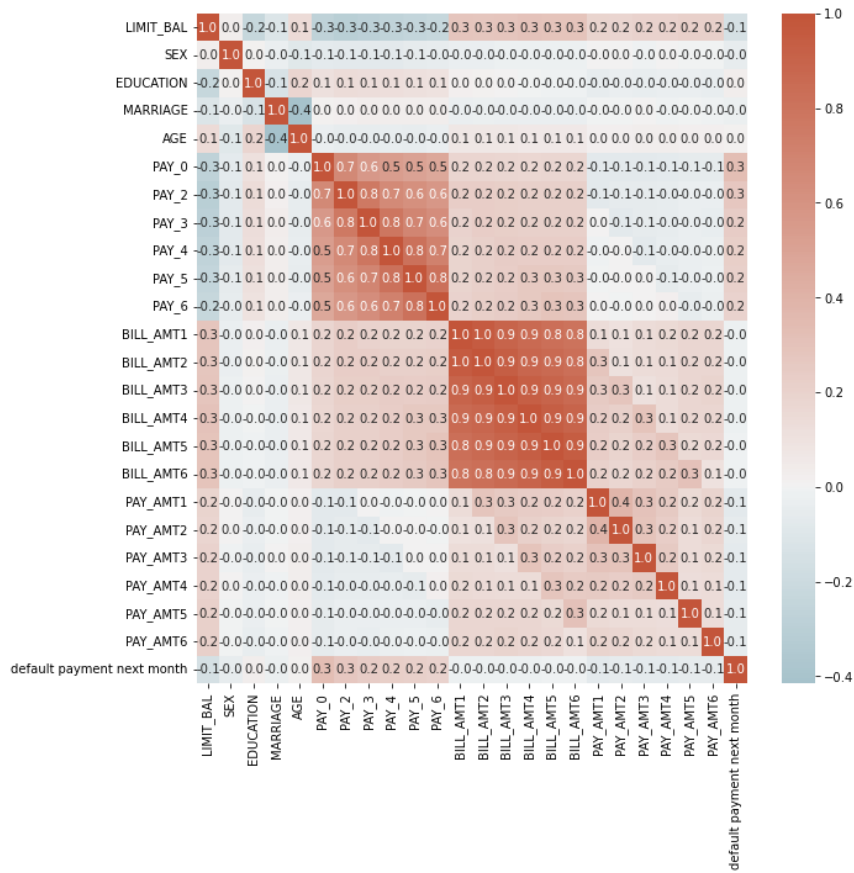

Figure 2. Correlation heatmap of Taiwan credit dataset

We compare the robustness of three models trained on this dataset: Generalized Linear Model (GLM), eXtreme Gradient Boosting (XGB), and Feed Forward Neural Network (FFNN). We tuned the hyperparameters to give optimal performance on a validation set. The performance metrics of these models are given in Table 5-1. The variable importance of each of these models are provided in Table 8-3 in the Appendix. We observed that GLM and XGB have the same top three important variables: 'PAY_0', 'BILL_AMT1' and 'LIMIT_BAL'. Whereas FFNN has high importance on the non-numeric variables compared to GLM and XGB. Also, XGB places more importance on 'PAY_AMT' variables compared to GLM and FFNN.

Table 5-1. Model Performance metrics

| Model | Metric: Log-loss | | | Metric: AUC | | |
|---|---|---|---|---|---|---|
| | Train | Test | Gap | Train | Test | Gap |
| GLM | 0.447 | 0.456 | 0.009 | 0.753 | 0.745 | 0.008 |
| XGB | 0.406 | 0.433 | 0.027 | 0.813 | 0.778 | 0.035 |
| FFNN | 0.421 | 0.442 | 0.021 | 0.792 | 0.765 | 0.027 |

From Table 5-1, we observe that the XGB model is performing best on the test data, but it also has the highest gap between training and test set performance. The gap statistic is a measure of the degree of overfitting, and a higher gap usually indicates that the model is unable to generalize well on unseen data. We will run robustness tests with ArPPV metrics to confirm whether the larger gap is cause for concern in this case.

We perturb the variables in the test dataset using the raw and adaptive methodology described in section 3, using the settings described in Table 5-2, with K=100 perturbations for each observation. For discrete variables, the noise scale was artificially raised to allow perturbations after rounding to the nearest integer. Categorical data needs a high budget for the same reason. As discussed in section 3, the perturbations are independent for numeric and non-numeric (categorical) variables, and especially for this dataset there is no reason to assume any association between the categorical and numeric variables because the categorical variables SEX, EDUCATION, and MARRIAGE are subjectively unrelated to numeric variables which is also evident from the correlation heatmap shown in Figure 2.

**Table 5-2. Perturbation settings**

| Variable type | Variables | Budget |
| --- | --- | --- |
| Categorical | 'SEX', 'EDUCATION', 'MARRIAGE' | 0.2 |
| Numeric | Rest of the variables | 0.05 |

We use different metrics to summarize the perturbations as discussed in section 2.2, including ArPPV. In Figure 3, we show the results from summarizing the perturbations using different metrics like absolute mean, absolute maximum, and root mean square (ArPPV). The bars in the plot corresponding to root mean square (rms) gives us the ArPPV. These metrics show mostly consistent ranking of models. Based on these summaries, for raw perturbations we observe that GLM is the most robust model for most metrics, followed by XGB and then FFNN. In the case of adaptive perturbations, we find GLM and XGB to be comparable according to absolute mean and root mean square, whereas the FFNN model is relatively more volatile.

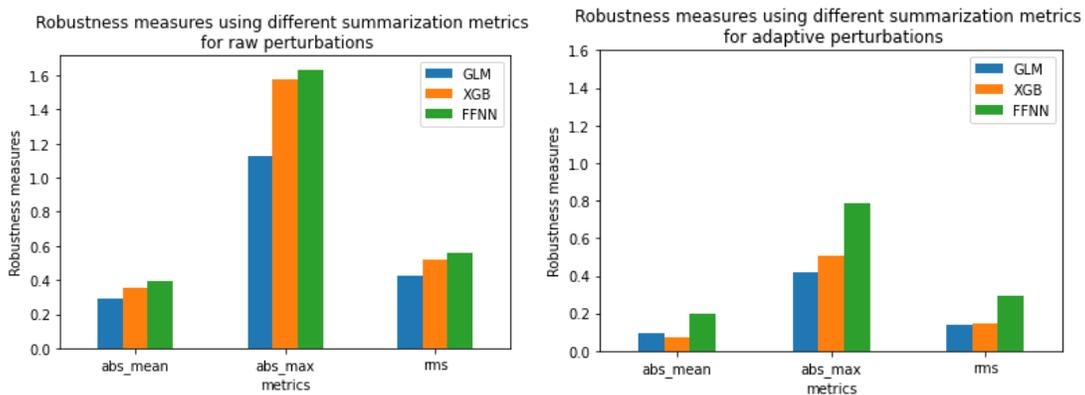

**Figure 3. Robustness measure from raw and adaptive perturbations computed using different summarization metrics for GLM, XGB and FFNN models for Taiwan credit dataset.**

In case of absolute maximum, we are not averaging the deviations across all perturbations but only considering the one perturbation which resulted in the maximum deviance from original prediction, naturally the values using this metric is higher than the rest and focuses on a more extreme scenario compared to the other metrics. Given that root mean square and absolute mean convey similar information in terms of ranking of the models, we have fixed the summarization metric as root mean square, using ArPPV as the robustness measure.

We also compute AUC for each set of perturbations using the true response values and compare with original data as shown in Figure 4. Similar comparison can be done using other metrics as well. These measures give us the performance of the models on the perturbed data, in contrast to the ArPPV, which isolates the stability of the model predictions to small perturbations.

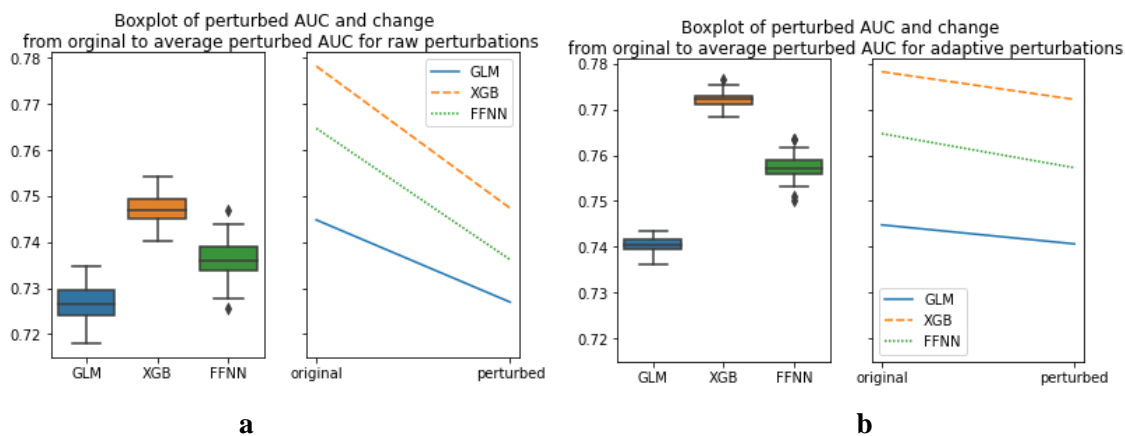

a  b

**Figure 4. Comparison of model performance on original and perturbed data of Taiwan credit dataset.**
  a) Boxplot of perturbed AUC score on 100 raw perturbations and change of AUC from original to perturbed data where the AUC for perturbed data is averaged across 100 raw perturbations.
  b) Boxplot of perturbed AUC on 100 adaptive perturbations and change of AUC from original to perturbed data where the AUC for perturbed data is averaged across 100 adaptive perturbations.

Figure 4a shows the decrease of average perturbed AUC from the original AUC for raw perturbations. Overall, XGB model still outperforms the other two models on the perturbed data and has the smallest variance on perturbed AUC as observed from the box plot. The decrease in performance for XGB and FFNN are sharper than GLM. Figure 4b conveys a similar information for the adaptive perturbations. XGB and GLM are similar in their AUC variance. XGB still retains higher performance in perturbed data. The analysis till this stage indicates that in spite of the larger gap, XGB model does not show significant higher volatility to perturbations in comparison to the other models.

## 5.1 Effect of budget on ArPPV

The perturbation strategies for numeric and non-numeric variables both include a measure of locality (called *budget*) to control the extent of perturbations of the variables. This is an important hyperparameter associated with generating perturbations since the robustness of models is tested for small-scale perturbations only. We do not want to test stability at very high budgets, as the model output is expected to change with significant changes in covariates. By its very nature, the model is expected to capture the response's relationship with its covariates, and a high-performing model will respond to significant change in its predictors. Hence all robustness tests need to be performed at low budgets.

We have separate strategies for perturbing numeric vs non-numeric variables. In this section, we isolate their impact in our illustrations. Consequently, we demonstrate the effect of the budget when perturbing numeric and categorical variables first separately and then simultaneously. First, we observed the effect of numeric perturbations against increasing budgets by observing the trend in ArPPV in Figure 5.

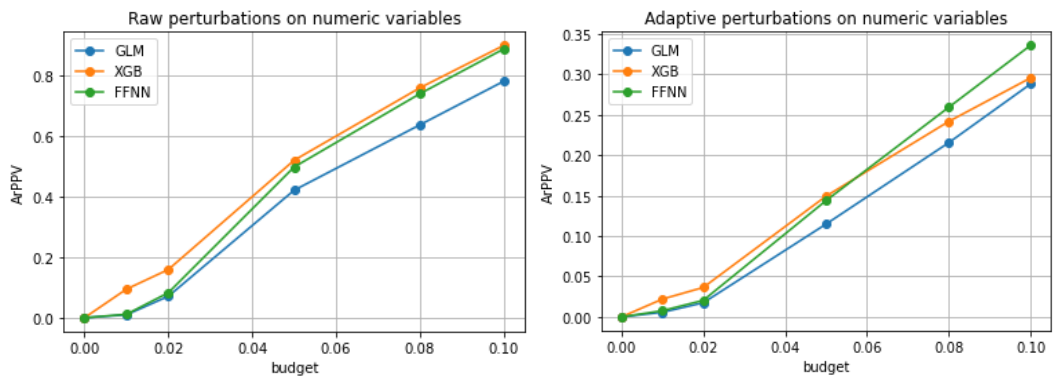

**Figure 5. Comparison of robustness of models to increasing perturbations on numeric variables.
(Left): Raw perturbations, (Right): Adaptive perturbations.**

We observe that, as expected, the ArPPV increases with increasing budget. For lower budgets, both the GLM and FFNN have lower ArPPV than the XGB. This is consistent with gap statistics given in Table 5-1. However, as budget increase we see ArPPV of FFNN rising at a greater rate than the other models. We also observe that as budget increases XGB model becomes similar to GLM model in terms of stability in case of adaptive perturbations. However, GLM continues to be the most stable model even at higher budgets. Although at 5% budget we see FFNN and XGB having similar volatility in Figure 5, this is in contrast with earlier result in Figure 3 which shows that FFNN model has a higher ArPPV. This is because the earlier result carries impact of both numeric and non-numeric variables perturbation. This will be further illustrated in Figure 6.

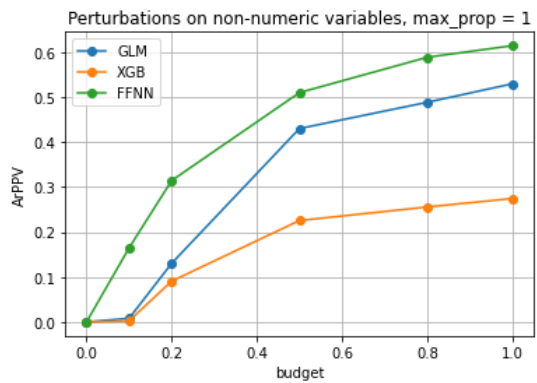

**Figure 6. Comparison of robustness of models to increasing perturbation on non-numeric variables.**

In Figure 6, we demonstrate the effect of budget on ArPPV for perturbation of categorical variables (Table 5-2). As before, ArPPV increases with increasing budget as expected, with the difference that the XGB model consistently has the lowest ArPPV followed by GLM and then FFNN. Also, XGB and GLM are more stable at lower budgets than FFNN. The high ArPPV of FFNN corresponding to categorical perturbations can once again be explained by observing the feature importance plots given in Table 8-3. The FFNN model gives more importance to categorical variables than the GLM and XGB models, which makes it more sensitive to perturbations in these variables. This behavior of FFNN explains why we observed high ArPPV for FFNN in Figure 3 but not when we were only perturbing numeric variables.

Finally, we perturb all the variables simultaneously and observe the change in ArPPV with increasing budgets. The results are shown in Figure 7. The x-axis shows the budget on numeric variables, whereas the budget of categorical variables is 5 times that budget.

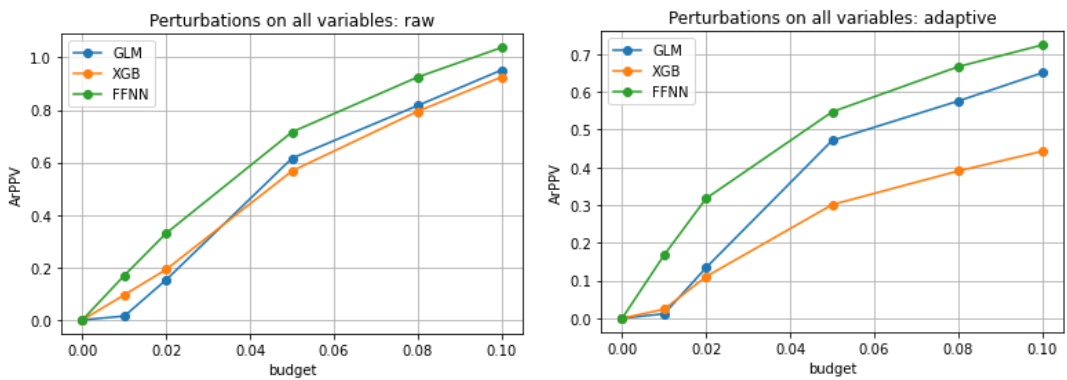

**Figure 7. Comparison of robustness of models to increasing perturbations.**
**(Left): Raw perturbations on numeric variables, (Right): Adaptive perturbations on numeric variables.**

Among the three models, FFNN remains the model with the highest ArPPV for the entire range of budgets. This is mostly because FFNN was more unstable to categorical perturbations than XGB or GLM. We observe that for small budget perturbations, both XGB and GLM have low ArPPV, although as budget increases the effect of categorical variables takes over and XGB becomes more stable than GLM, and this is more pronounced in adaptive perturbations. Due to GLM's sub-optimal performance on test dataset (Table 5-1), XGB is the preferred model based on both its performance on unseen data and robustness to small-scale perturbations.

## 5.2 Correlated and independent perturbations for numeric variables

The proposed methodology in section 3.1.1 generates perturbations that respect the inherent correlation in the data. In this section, we compare the correlated perturbations with perturbations that are generated independently on each variable. Here, we use a 5% budget for the numeric variables. The ArPPV results are given in Table 5-3. Both correlated and independent perturbations result in similar ordering of ArPPV for the models, with the GLM having the least ArPPV followed by FFNN and XGB.

**Table 5-3. Robustness measure ArPPV for correlated and independent numeric perturbations.**

| Model | ArPPV | | | |
| --- | --- | --- | --- | --- |
|  | *Raw* | | *Adaptive* | |
|  | *Correlated* | *Independent* | *Correlated* | *Independent* |
| GLM | 0.422 | 0.378 | 0.118 | 0.107 |
| XGB | 0.521 | 0.456 | 0.159 | 0.152 |
| FFNN | 0.499 | 0.426 | 0.149 | 0.139 |

We observe that the independent perturbations result in smaller ArPPVs compared to correlated perturbations, even though the rank ordering between the models remains the same. This is because of the highly correlated variables as seen in Figure 2; the importance of a given variable is shared with the variables which are highly correlated with it. Thus, correlated perturbations cause the prediction to change more in this dataset, since all the variables in the same group are perturbed in the same direction, leading to higher a ArPPV. On the other hand, for independent perturbations even though one of the variables changes, the other correlated variables may have not changed in the same direction. Sometimes these changes can be in opposing directions, and this can reduce the impact of the resulting perturbations.

Table 5-4. Comparing the correlation strength of raw and adaptive perturbations generated from correlated and independent method for a single observation and correlated variables.
'BILL_AMT6' vs 'BILL_AMT5'

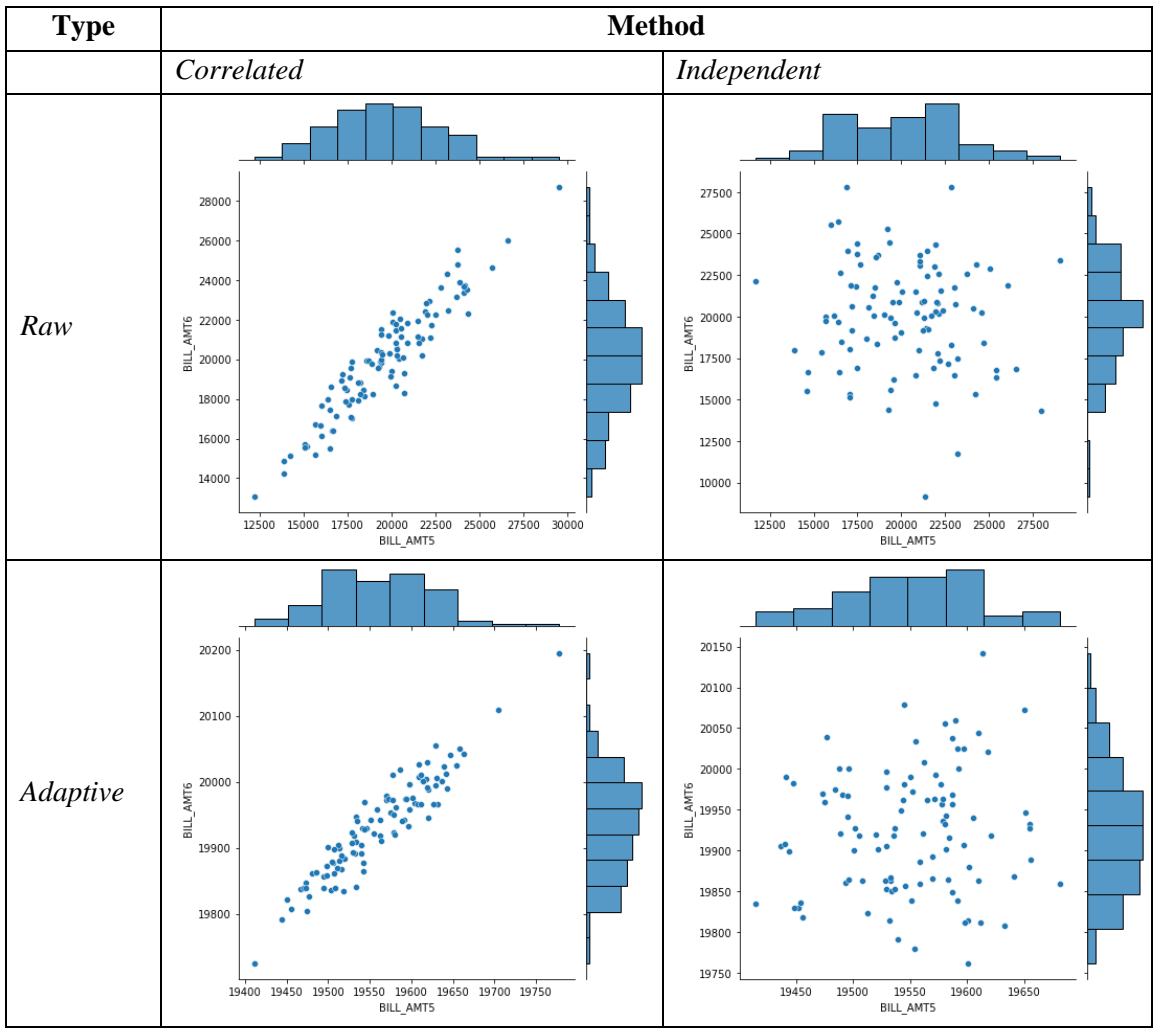

In Table 5-4, we show the perturbed values for a single observation and a scatter plot of correlated variables 'BILL_AMT5' and 'BILL_AMT6'. These variables' actual value for the observation is 19549 and 19920, respectively. We observe that the perturbations generated from correlated method maintain the correlation between the two variables around the actual point, whereas the perturbations from independent methods are randomly scattered around the actual data point, thus do not maintain the correlation. A similar comparison is given for variables 'LIMIT_BAL' and 'PAY_AMT2' in appendix Table 8-2 whose actual values for given observation are 20000 and 1405 respectively. These two variables do not have a strong correlation with each other, so perturbations from both methods are randomly scattered around the actual data point.

Note that for small scale perturbations the difference in impact between correlated and independent perturbations is typically not large. However, in the rare cases where highly correlated variables are included as model predictors the difference can still be significant as seen in Table 5-3.

## 5.3 Comparison of 'Pseudo-distance' method and 'shuffling' method for categorical variables

In this section we compare the pseudo-distance and shuffling methods for perturbing categorical variables in the Taiwan credit data. As introduced in section 3.2.1, the pseudo-distance measures the difference between two levels of a categorical variable based on their average impact on the response. This means that for a given dataset, any two levels of a categorical variable are considered similar by this metric if they have a similar average impact on the response and perturbing that variable between these two levels should have a lower impact than perturbing them to a level which has a very different average response.

**Table 5-5. Average response for each level of a categorical variables 'EDUCATION' and 'MARRIAGE' defining the similarity among the levels.**

| EDUCATION | |
|---|---|
| *Variable level* | *Average response* |
| graduate school | 0.197065 |
| university | 0.234813 |
| high school | 0.256193 |
| others | 0.076655 |

| MARRIAGE | |
|---|---|
| *Variable level* | *Average response* |
| married | 0.234975 |
| single | 0.211688 |
| others | 0.235808 |

For example, we can observe from Table 5-5 that for the variable 'MARRIAGE', levels 'married' and 'others' are similar to each other based on their average impact on response. Here 'others' may include partners, divorced, separated, etc. Similarly, for 'EDUCATION', we observe that average proportion of default decreases from 'high school' to a 'university degree' (undergraduate), while the default rate is still lower for people with a graduate degree. According to Table 5-5, a university education and high school education are considered more similar than a university education and graduate degree. Thus, perturbing this variable from 'university' to 'high school' or 'high school' to 'university' will result in a perturbation which is small-scale and local as compared to when we perturb 'university' or 'high school' to 'others'. Note that using this method is only approximating closeness of the levels artificially and may not be representative of a more subject-matter-based notion of closeness of the levels.

The pseudo-distance method of categorical perturbations ensures that the perturbations occur within similar levels of a variable as compared to the shuffling method where the perturbations are generated by randomly shuffling the value of a variable using its marginal distribution. To further illustrate the difference between the two methods, we choose a few observations given in Table 5-6 and plot the distribution of the perturbed values under the two methods. The perturbation budget for the pseudo-distance method is set at 40% and we have plotted the results for two different settings of *max_prop*, namely max_prop=1 indicated by 'pseudo' and max_prop=0.5 indicated by 'pseudo - 0.5'. As discussed in section 3.2.2, $max\_prop$ indicates the proportion of $K$=100 perturbations to be accepted for an observation and helps in creating more controlled perturbations especially in extreme cases of binary variables. The title in each plot indicates the actual value of the variable for that observation.

**Table 5-6. Distribution of perturbations of categorical variables under pseudo-distance and shuffling perturbation strategies.**

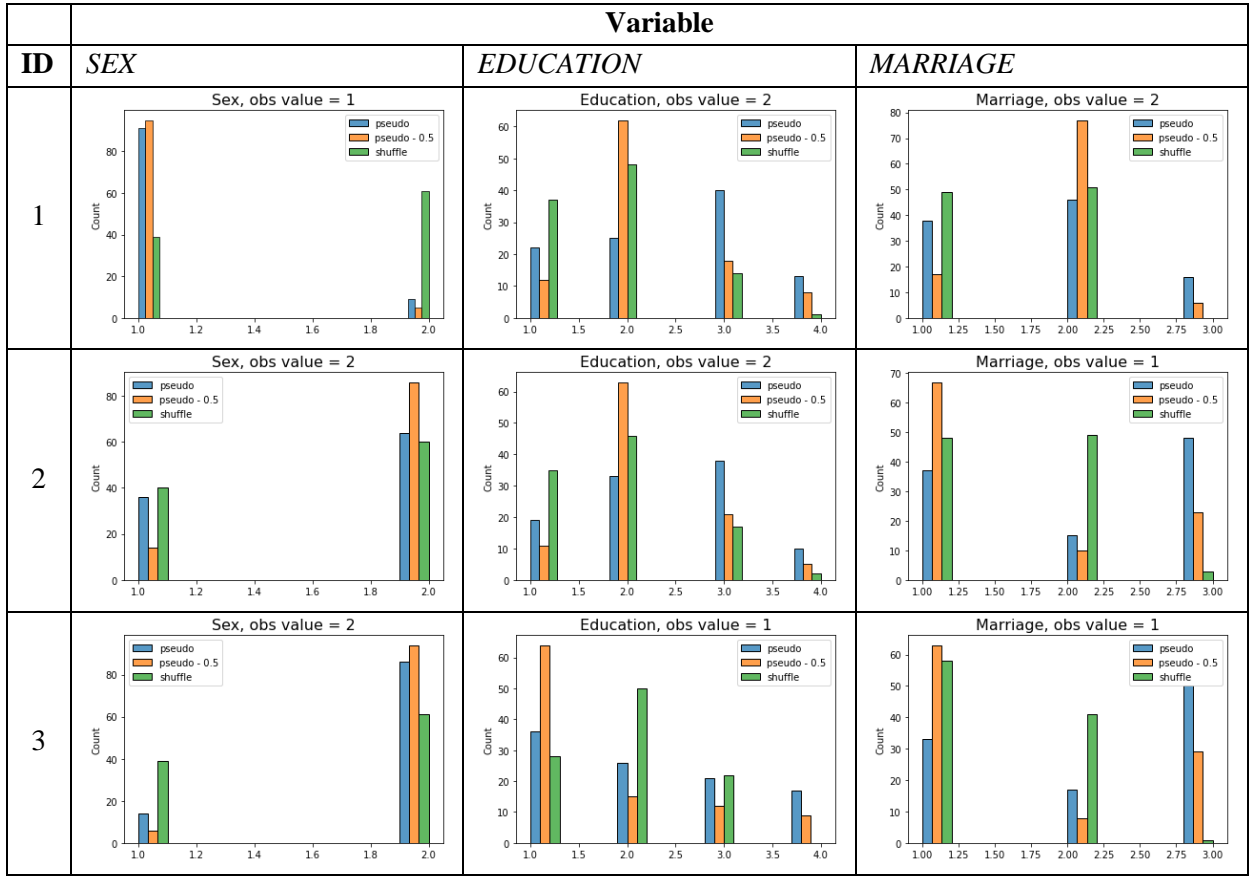

It can be observed from the plots in Table 5-6 that under the pseudo-distance strategy with max_prop=0.5, about 50% of the time an observation is unperturbed, and when it is perturbed, it prefers movement to levels that are similar to its actual level over the others. For example, the actual level (value) of 'EDUCATION' is 2 (university) for observation id 1, where we can observe that more than 50% perturbations lie at level 2 for 'pseudo – 0.5', and it tends to move to level 3 (high school) more than any other level because level 2 has the highest similarity with level 3 as shown in Table 5-5. Similarly for observation id 2, the actual level of 'MARRIAGE' is 1 (married), the 'pseudo – 0.5' strategy does not easily perturb the point, and when it indeed perturbs the point, it prefers the level with maximum similarity to level 1, i.e., level 3 (others). The 'pseudo' method accepts all the generated perturbations of an observation resulting in more perturbations compared to 'pseudo - 0.5' method.

On the other hand, the 'shuffle' method just perturbs the variable following its marginal distribution, irrespective of the actual observation. Hence, we observe that the perturbation distribution of all 3 observations is identical for the 'shuffle' method.

## 5.4 Local diagnosis on Taiwan credit dataset

It is observed in Figure 5 that XGB has the higher volatility at low (<5%) budget for both raw and adaptive perturbations for numeric variables. We want to identify the source of this higher volatility in this model. We conduct local diagnosis (as described in section 4) of the XGB model to understand its local behavior and identify the regions of data where it is least stable. We will do this analysis for the numeric variables using 2% perturbation budget. The local analysis consists of four parts, we first carry out the PSI (Figure 8) and diagnostic partition tree (Figure 9) analysis for 10% worst observations when simultaneously perturbing all variables. This allows us to find variables which leads to maximum volatility in the XGB model. Next, we perturb one variable at a time and compute ArPPV for all three models (Figure 10) which helps us identify regions where the XGB model shows more volatility for a particular variable compared to other models. Thus, identifying key variables of interest, we follow up on these variables using the single variable diagnosis mentioned in Section 4.

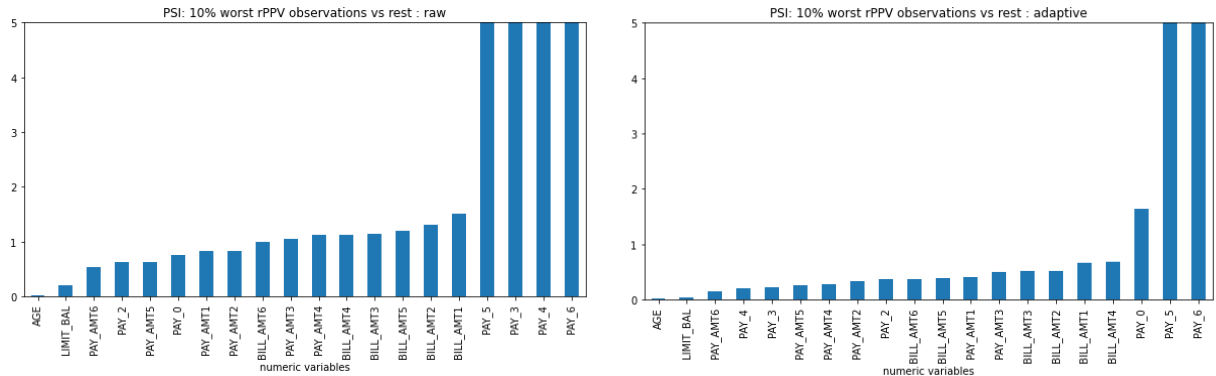

**Figure 8. PSI based on observations with worst 10% rPPV for XGB model.**
**(Left: raw, Right: adaptive)**

Based on Figure 8 and Figure 9, 'BILL_AMT1' is the prime source of volatility at 2% budget for raw perturbation, whereas 'PAY_0' is the top variable of interest for adaptive perturbations followed by 'BILL_AMT4'. The diagnostic trees also split on some of the 'PAY_AMT' variables, although they are not the first splits indicating these variables as a source of volatility as well. PSI for some of the PAY variables is infinite (Table 8-4), suggesting a difference in support of the distributions of the worst observations vs rest as demonstrated in Table 8-5 for raw perturbations. Note that the PAY, BILL_AMT and PAY_AMT variables are highly correlated and hence we might see a raised PSI due to this correlation or a correlated variable being a proxy for the split on another variable.

The variable-by-variable analysis confirms 'BILL_AMT1' as not only a high source of volatility for the XGB model but also a variable where the XGB model's volatility is significantly higher than the other models for both raw and adaptive perturbations. The 'PAY_AMT' variables are also identified as sources of added volatility specifically for XGB models and this is supported by their higher variable importance for the XGB model (Table 8-3). Among the 'PAY_AMT' variables, 'PAY_AMT2' ranks high in both raw and adaptive perturbation and is also a split variable for the raw perturbation diagnostic tree in Figure 9. Hence, we shall do a follow-up analysis on the two variables 'BILL_AMT1' and 'PAY_AMT2'. Note that the ArPPV for adaptive perturbations in Figure 10 are much smaller than their raw counterparts.

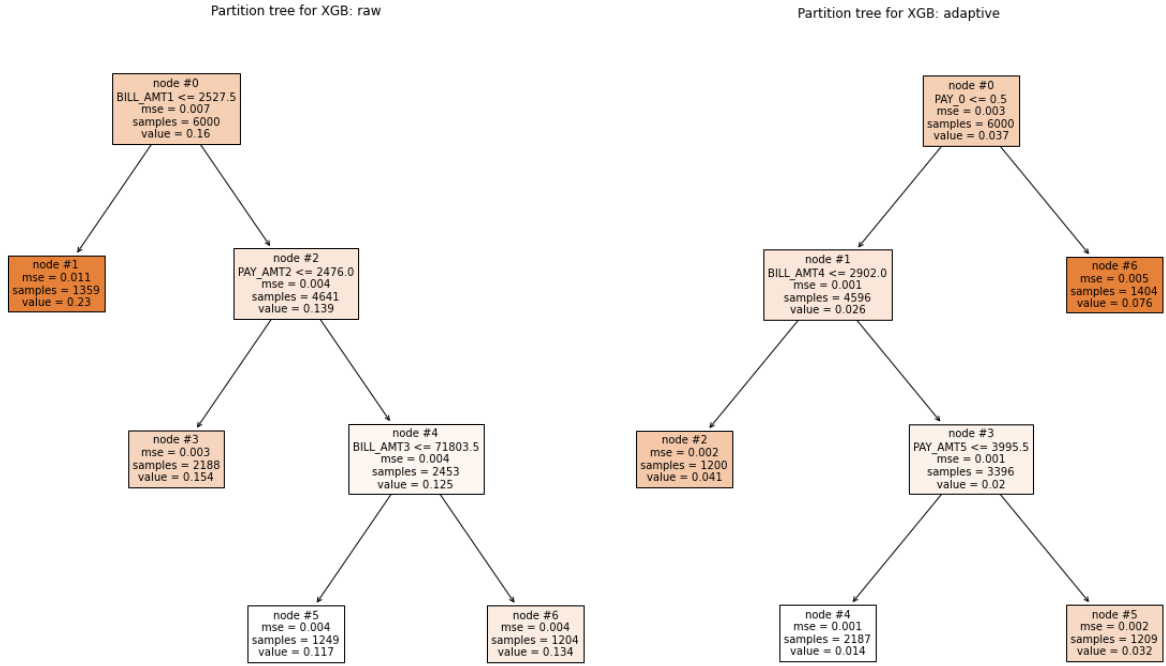

**Figure 9. Supervised diagnostic partition tree developed using rPPV of XGB model as response. (Left: raw, Right: adaptive)**

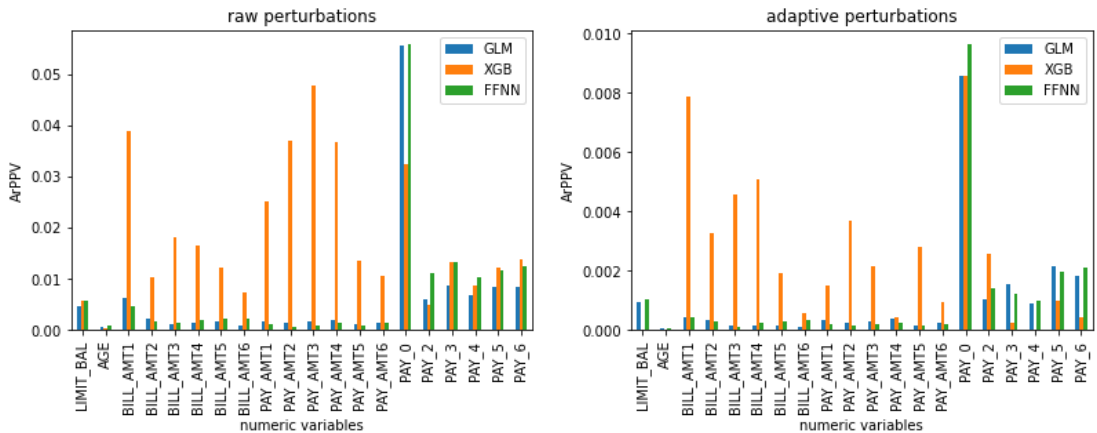

**Figure 10. ArPPV comparison for single variable perturbations.**

In the follow up diagnosis, we perturb the variable of interest and compute the rPPV for each observation with 2% budget. The results for 'BILL_AMT1' are given in Table 5-7 and that of 'PAY_AMT2' in Table 5-8.

The diagnosis for 'BILL_AMT1' shows that XGB model has a more jagged PDP than the other models resulting in a number of monotone violations. Although the FFNN PDP captures similar model behavior, it has a more stable behavior resulting in less (or no) monotone violations. We can see that the histogram of 'BILL_AMT1' is left skewed with high data concentration on the lower values. The adaptive perturbations thus produce more conservative perturbations in this range. However, even with these smaller perturbations we observe that the XGM model is more volatile in this range than the other two models.

**Table 5-7. Single variable diagnosis of GLM, XGB and FFNN models on 'BILL_AMT1' variable.**

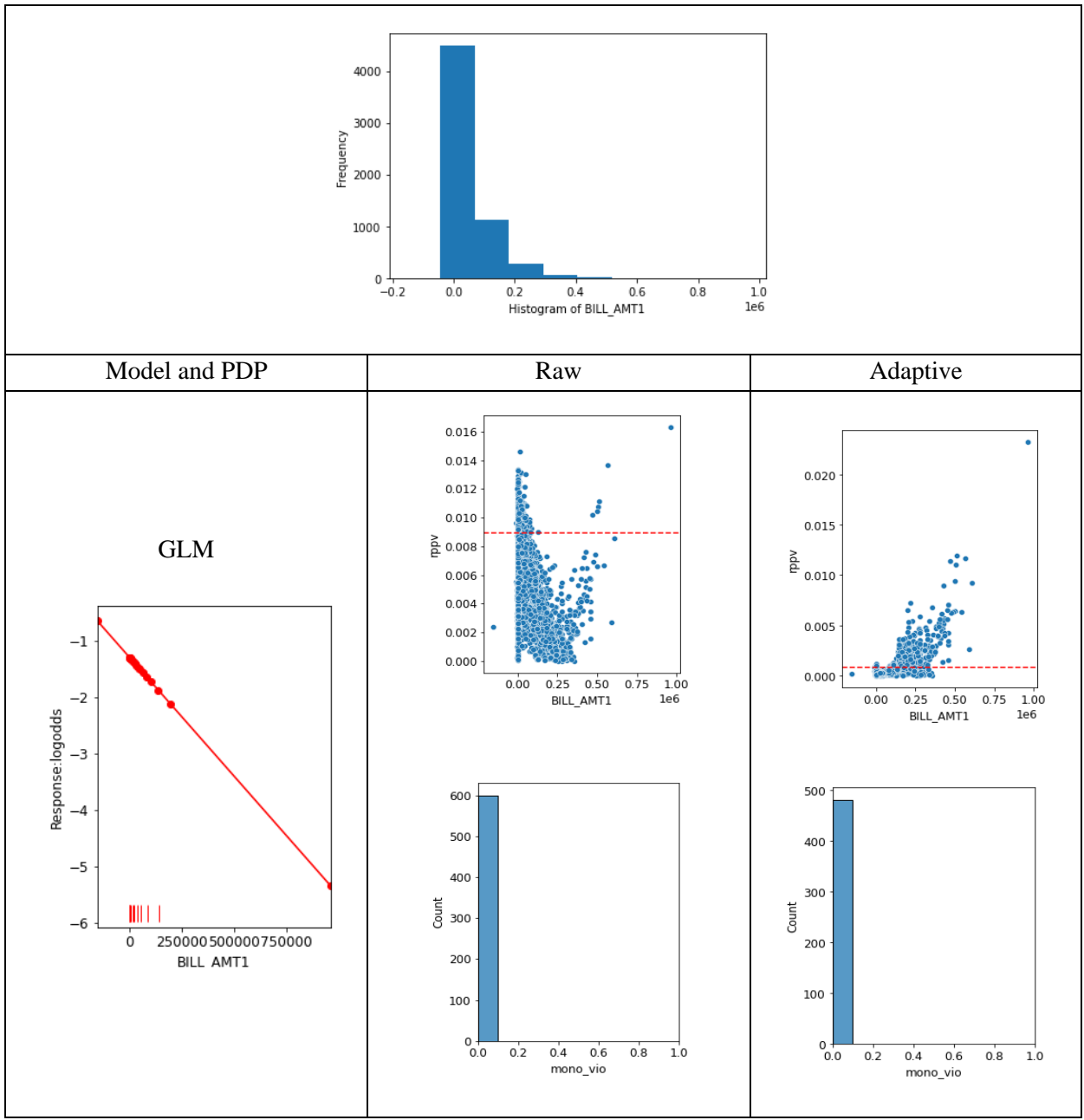

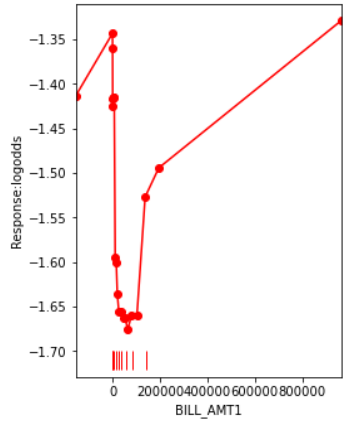
XGB
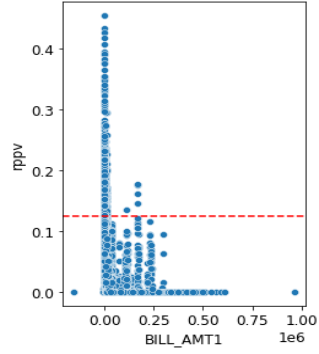
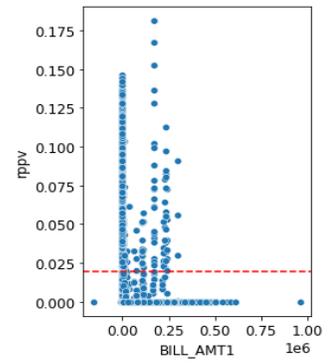
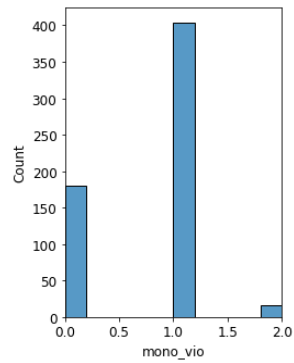
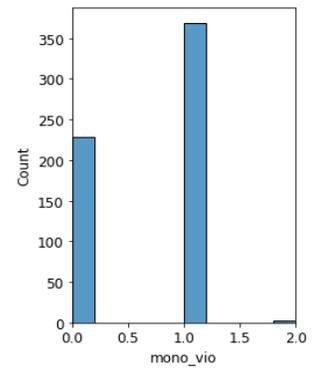
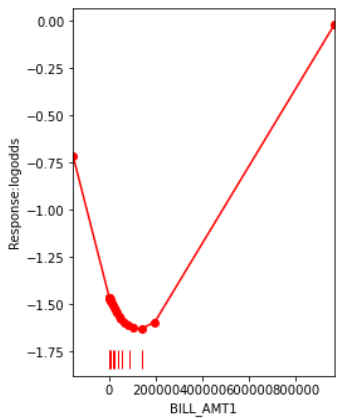
FFNN
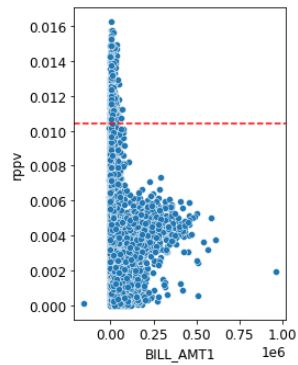
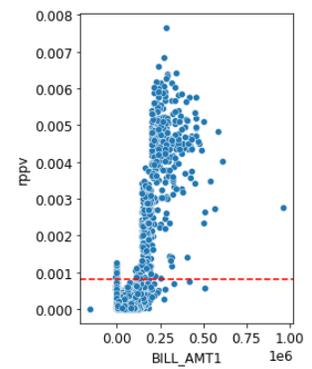
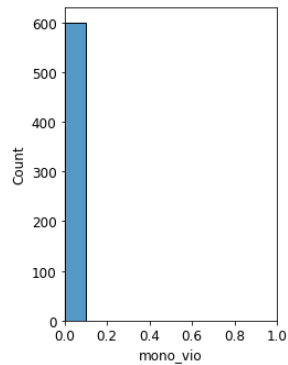
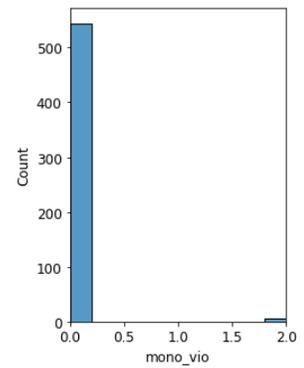

The PDP of the XGB model for 'PAY_AMT2' once again shows a sharp transition in Table 5-8 which causes the high rPPV values in this model compared to the others. For the raw perturbations, the high rPPV (>0.1) points also have a substantial number of monotone violations showing that the transition is not smooth and the raised rPPV values occur due to a combination of sensitivity and lack of robustness. Under the adaptive perturbations, we have lowered the perturbation strength in this transition region which has significantly reduced the number of monotone violations although the observations in the lower range of this variable still have higher rPPV for XGB model than the other models. Note that FFNN models also show some monotone violations with adaptive perturbations, possibly caused by the higher perturbations in the sparse regions. However, the rPPV for these observations are less than 0.008 and hence these are not strong enough violations to cause concern.

**Table 5-8. Single variable diagnosis of GLM, XGB and FFNN models on 'PAY_AMT2' variable.**

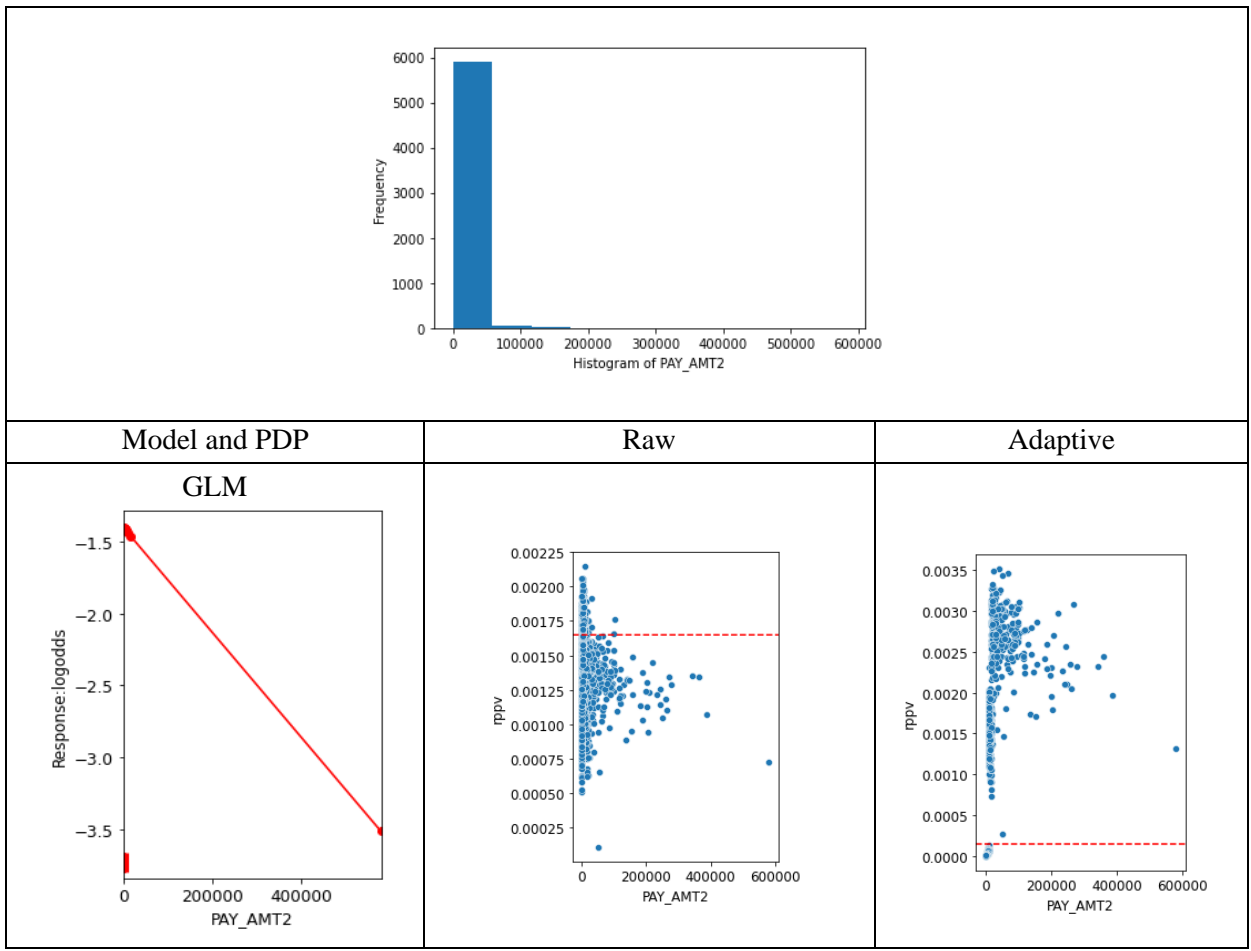

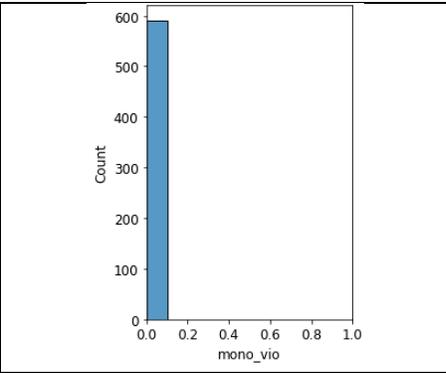
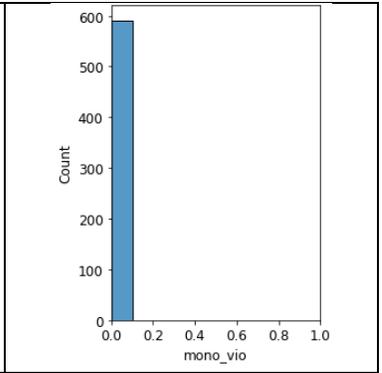
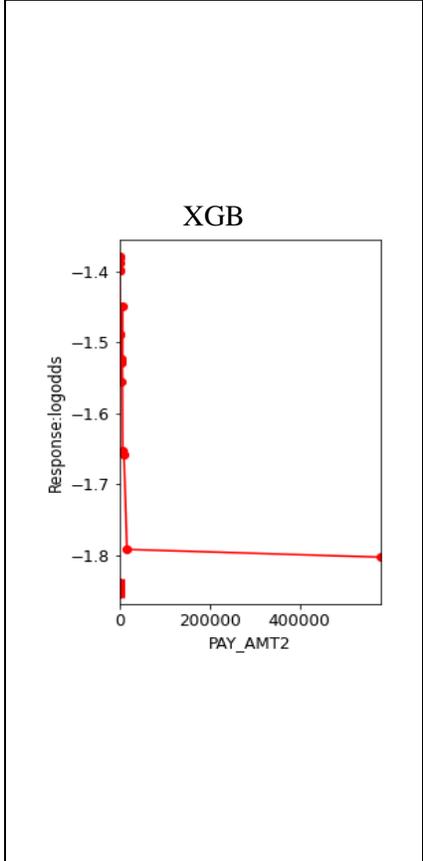
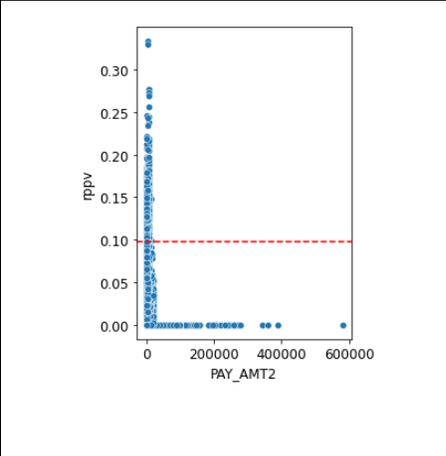
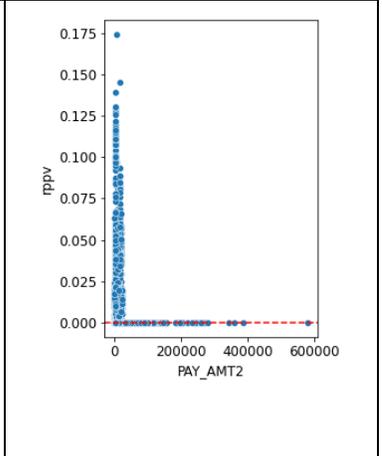
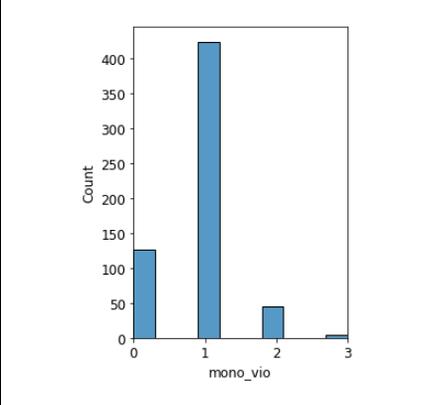
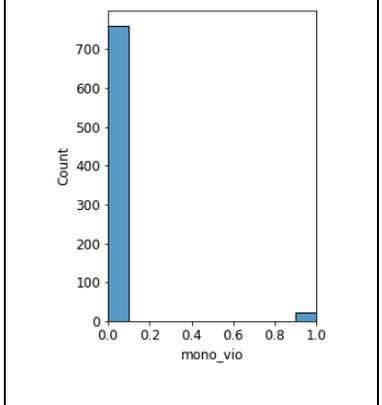
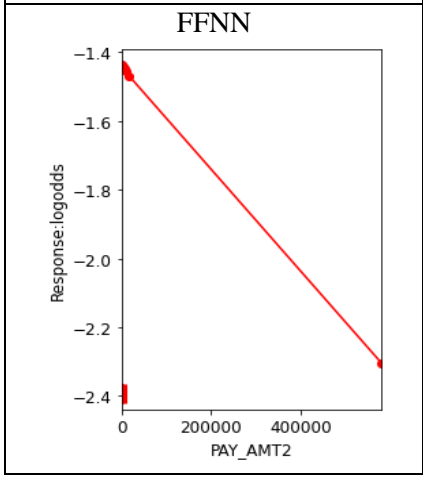
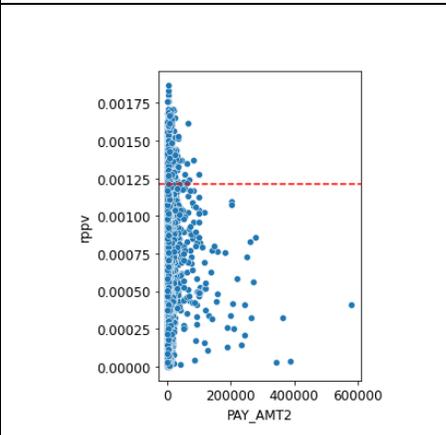
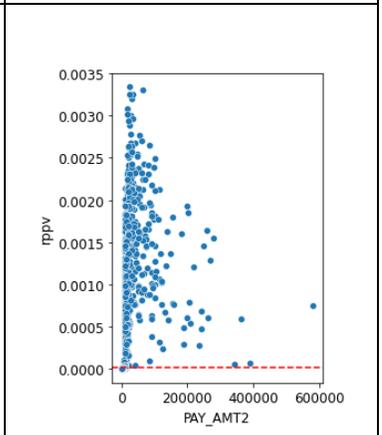

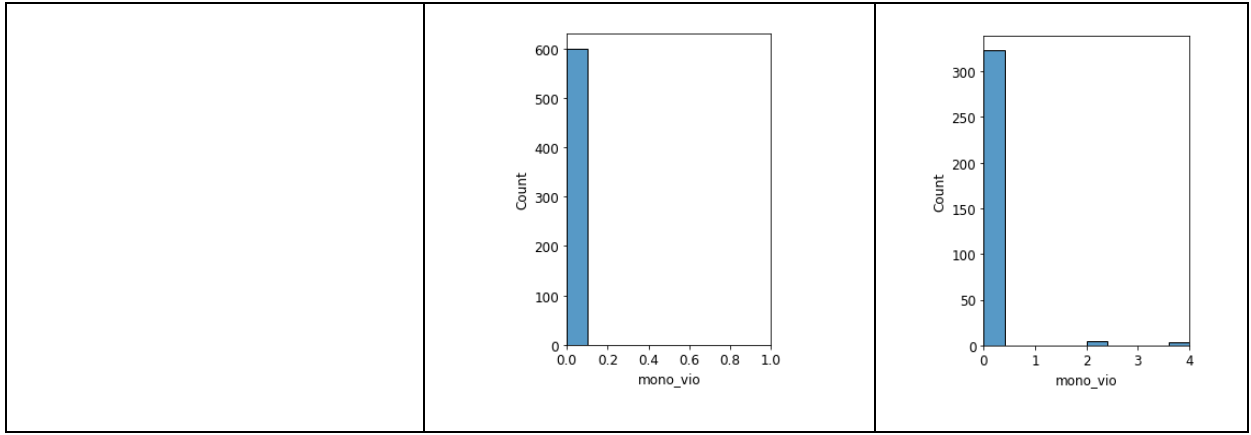

The analysis has helped us understand which variables are causing the XGB model to be more volatile than the others at low budget which may account for the slightly higher gap in this model. As seen from the monotone violations, the robustness of the XGB model may be increased by adding some monotonicity constraints if applicable. However, given the overall behavior of the ArPPV curves in Figure 5 and Figure 7, XGB is not substantially non-robust to perturbations compared to the other models and has superior performance compared to the other models. On the other hand, FFNN models seem to deteriorate faster to increasing budgets and is specifically sensitive to the non-numeric covariates. Due to these issues, the XGB model is preferred over the FFNN model.

## 6  Conclusion

We introduced covariate perturbation as a methodology to assess model robustness as a model's ability to maintain stable outputs against small-scale random perturbations on the inputs by proposing tailored perturbation strategies for numeric (continuous and discrete) and non-numeric (categorical) variables. We introduced a robustness measure ArPPV to assess the impact of these perturbations to compare the robustness of multiple models. The adaptive distribution-based perturbation strategy for numeric variables ensures that the correlation structure of the original data is maintained in the perturbed data, and the notion of local perturbations is adhered in non-numeric variables by introducing a data-based joint perturbation strategy that relies on a concept of pseudo-distance measure. We further described the observation level measures of robustness that can be used to understand the behavior of a model locally and identify regions of the data where a model is particularly unstable.

We illustrated the impact of budget on these perturbations using the Taiwan credit dataset, and the change in predictions observed by the ArPPV measure. We also emphasized the need to perform the robustness test on small budgets to not conflate desired model sensitivity with robustness. Further, we introduced and compared multiple metrics to summarize the perturbations at an observation level that can be used in different contexts. The comparison of correlated and independent perturbations for numeric variables showed us that for smaller budgets, both methods are similar and for larger budgets, the correlated method appropriately maintains the relationships among variables. Similarly, the pseudo-distance method for categorical variables ensures conservative local perturbations compared to random shuffling of levels within a variable. Overall, this paper introduces techniques to compare multiple models on the basis of their robustness by generating small scale perturbations in the predictor space, and diagnostics to further investigate the perturbation results and isolate regions of the data requiring additional attention.

# 8 Appendix

This appendix contains additional analysis results from the Taiwan credit dataset analysis described in section 5.

**Table 8-1. Attribute information**

| Name | Description | Type | Remarks |
|---|---|---|---|
| LIMIT_BAL | Amount of the given credit | Continuous | In NT Dollar |
| SEX | Gender | Categorical | 1 = male<br>2 = female |
| EDUCATION | Education background | Categorical | 1 = graduate school<br>2 = university<br>3 = high school<br>4 = others |
| MARRIAGE | Marital status | Categorical | 1 = married<br>2 = single<br>3 = others |
| AGE | Age | Discrete | In year |
| PAY_0, PAY_2, PAY_3, PAY_4, PAY_5, PAY_6 | History of past payment (from April to September 2005) | Discrete | Ordered discrete variable in the range [-2, 8] |
| BILL_AMT1 – BILL_AMT6 | Amount of bill statement (from April to September 2005) | Continuous | In NT Dollar |
| PAY_AMT1 – PAY_AMT6 | Amount of previous payment (from April to September 2005) | Continuous | In NT Dollar |

**Table 8-2. Comparing the correlation strength of raw and adaptive perturbations generated from correlated and independent method for a single observation and uncorrelated variables.**
'PAY_AMT2' vs 'LIMIT_BAL'

| Type | Method | |
|---|---|---|
| | *Correlated* | *Independent* |
| *Raw* | 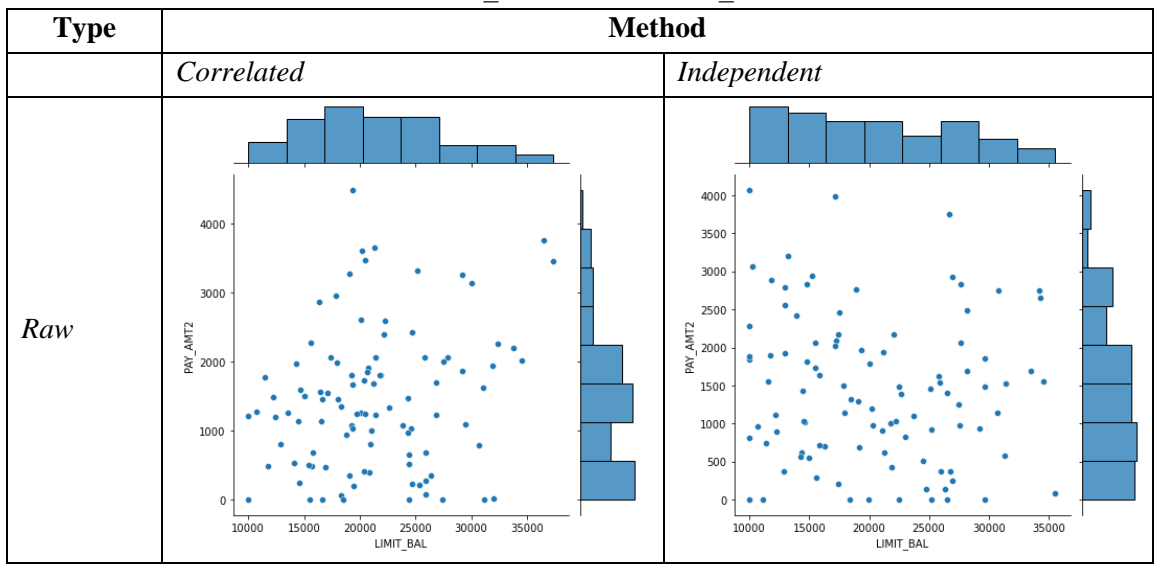 | |
| *Adaptive* | 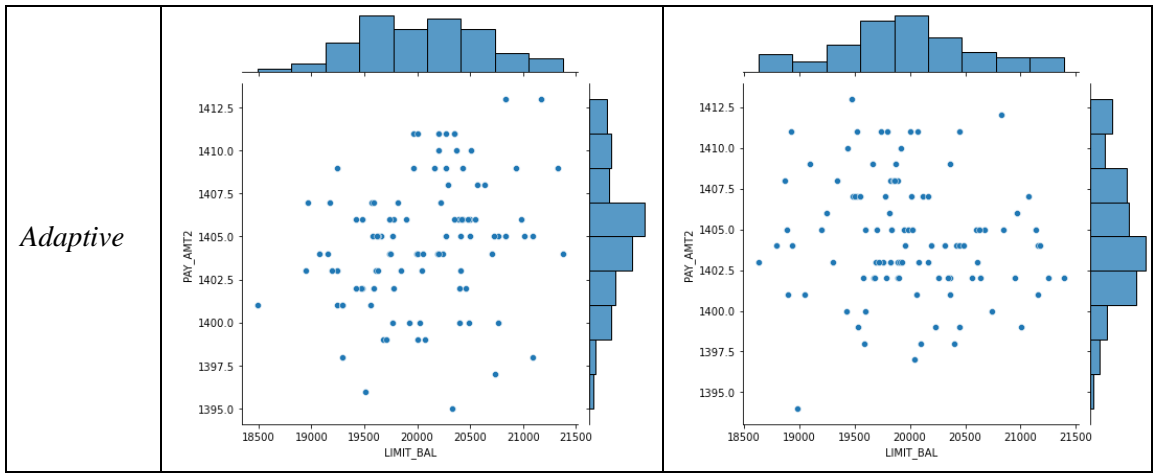 | |

**Table 8-3. Permutation-based feature importance of GLM, XGB and FFNN models.**

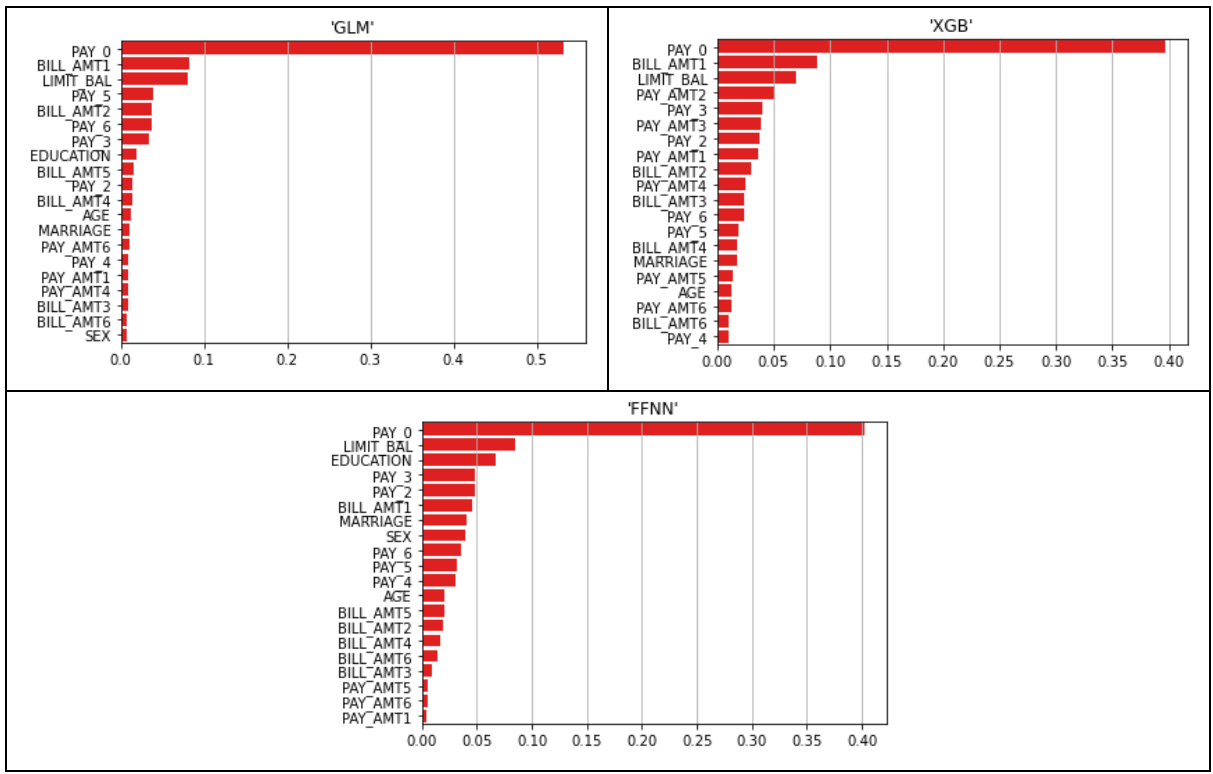

**Table 8-4. PSI: Calculated between 10% worst observation based on rPPV and rest of the observations.**

| | raw | | | adaptive | | |
|---|---|---|---|---|---|---|
| | **GLM** | **XGB** | **FFNN** | **GLM** | **XGB** | **FFNN** |
| **LIMIT_BAL** | 0.149 | 0.206 | 0.038 | 0.453 | 0.045 | 0.163 |
| **AGE** | 0.019 | 0.03 | 0.024 | 0.057 | 0.02 | 0.036 |
| **BILL_AMT1** | 2.013 | 1.508 | 0.77 | 0.27 | 0.657 | 0.105 |
| **BILL_AMT2** | 2.038 | 1.305 | 0.724 | 0.255 | 0.514 | 0.078 |
| **BILL_AMT3** | 2.043 | 1.152 | 0.726 | 0.282 | 0.513 | 0.056 |
| BILL_AMT4 | 1.985 | 1.134 | 0.703 | 0.294 | 0.69 | 0.087 |
| BILL_AMT5 | 1.903 | 1.191 | 0.713 | 0.363 | 0.381 | 0.112 |
| BILL_AMT6 | 1.649 | 0.996 | 0.684 | 0.34 | 0.375 | 0.112 |
| PAY_AMT1 | 0.782 | 0.834 | 0.265 | 0.164 | 0.401 | 0.131 |
| PAY_AMT2 | 0.776 | 0.834 | 0.418 | 0.082 | 0.331 | 0.072 |
| PAY_AMT3 | 0.662 | 1.057 | 0.372 | 0.082 | 0.499 | 0.069 |
| PAY_AMT4 | 0.633 | 1.117 | 0.197 | 0.026 | 0.286 | 0.062 |
| PAY_AMT5 | 0.53 | 0.637 | 0.213 | 0.033 | 0.258 | 0.033 |
| PAY_AMT6 | 0.466 | 0.535 | 0.157 | 0.064 | 0.15 | 0.055 |
| PAY_0 | inf | 0.765 | inf | 4.83 | 1.645 | 2.346 |
| PAY_2 | inf | 0.627 | inf | 2.037 | 0.366 | 0.668 |
| PAY_3 | 1.147 | inf | inf | 1.328 | 0.216 | 0.405 |
| PAY_4 | 1.051 | inf | inf | inf | 0.211 | 0.381 |
| PAY_5 | 1.269 | inf | inf | 1.251 | inf | inf |
| PAY_6 | inf | inf | inf | inf | inf | inf |

Table 8-5. Detailed PSI calculation table for 'PAY_6' variable of 'XGB' model.

| | Category | Base (% total) | New (% total) | ln (New/Base) | New-Base | Index |
|---|---|---|---|---|---|---|
| 1 | 7 | 0.0019 | 0.0033 | 0.5878 | 0.0015 | 0.0009 |
| 2 | 6 | 0.0007 | 0.0017 | 0.8109 | 0.0009 | 0.0008 |
| 3 | 5 | 0.0002 | 0.0000 | -inf | -0.0002 | inf |
| 4 | 4 | 0.0011 | 0.0033 | 1.0986 | 0.0022 | 0.0024 |
| 5 | 3 | 0.0061 | 0.0050 | -0.2007 | -0.0011 | 0.0002 |
| 6 | 2 | 0.0998 | 0.0483 | -0.7252 | -0.0515 | 0.0373 |
| 7 | 0 | 0.5098 | 0.8800 | 0.5459 | 0.3702 | 0.2021 |
| 8 | -1 | 0.2011 | 0.0350 | -1.7485 | -0.1661 | 0.2904 |
| 9 | -2 | 0.1793 | 0.0233 | -2.0390 | -0.1559 | 0.3179 |
| | | | | | PSI ($\sum index$) = | inf |

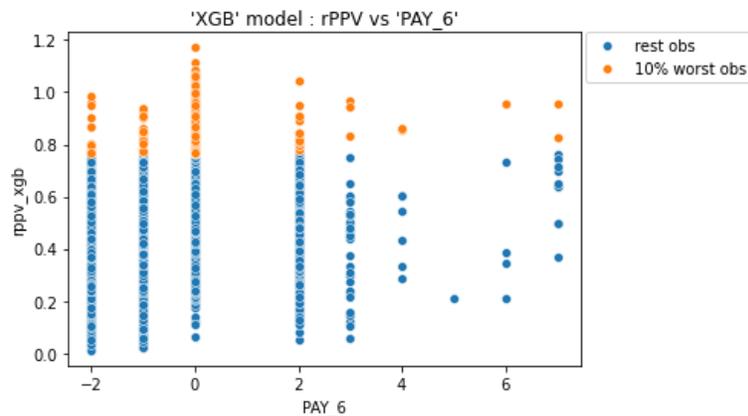

Figure 11. rPPV vs 'PAY_6' for XGB variable illustrating the difference in distribution of 10% worst observations and rest of the observations for raw perturbations.

We observe from the table that 'PAY_6' is a discrete variable with few unique values, some of which are absent in the worst-case distribution (Figure 11). The infinite PSI arises when any bins in target or reference distribution have zero observations, as shown in the case of 'PAY_6' variable of the XGB model in Table 8-5.